\documentclass[10pt,twocolumn,letterpaper]{article}

\usepackage{iccv}
\usepackage{times}
\usepackage{epsfig}
\usepackage{graphicx}
\usepackage{amsmath}
\usepackage{amssymb}
\usepackage{multirow}
\usepackage{balance}

\newcommand{\wraptxt}[1]{#1\xspace}
\newcommand{\figref}[1]{\wraptxt{Figure \ref{#1}}}

\newcommand{\tabref}[1]{\wraptxt{Table \ref{#1}}}
\newcommand{\pn}{\wraptxt{Prototypical Network}}
\newcommand{\pns}{\wraptxt{Prototypical Networks}}

\newcommand{\Pns}{\wraptxt{Prototypical Networks}}
\newcommand{\shot}[2]{#1-shot #2-way}

\newcommand{\mi}{\wraptxt{miniImagenet}}
\newcommand{\ti}{\wraptxt{tieredImagenet}}
\newcommand{\fs}{\wraptxt{CIFAR Few-Shot}}

\usepackage{floatrow}

\usepackage[pagebackref=true,breaklinks=true,letterpaper=true,colorlinks,bookmarks=false]{hyperref}

\iccvfinalcopy 

\def\iccvPaperID{3571} 

\ificcvfinal\fi

\begin{document}

\title{
Few-Shot Learning with Embedded Class Models and Shot-Free Meta Training}

\author{Avinash Ravichandran\\
Amazon Web Services\\
{\tt\small ravinash@amazon.com}
\and
Rahul Bhotika\\
Amazon Web Services\\
{\tt\small bhotikar@amazon.com}
\and
Stefano Soatto\\
Amazon Web Services and UCLA\\
{\tt\small soattos@amazon.com}
}

\maketitle

\newpage
\begin{abstract}
We propose a method for learning embeddings for few-shot learning that is suitable for use with any number of shots (shot-free). Rather than fixing the class prototypes to be the Euclidean average of sample embeddings, we allow them to live in a higher-dimensional space (embedded class models) and learn the prototypes along with the model parameters. The class representation function is defined implicitly, which allows us to deal with a variable number of shots per class with a simple constant-size architecture. The class embedding encompasses metric learning, that facilitates adding new classes without crowding the class representation space. Despite being general and not tuned to the benchmark, our approach achieves state-of-the-art performance on the standard few-shot benchmark datasets.
\end{abstract}

\begin{figure}[htb]
\begin{center}
\includegraphics[height=.30\linewidth]{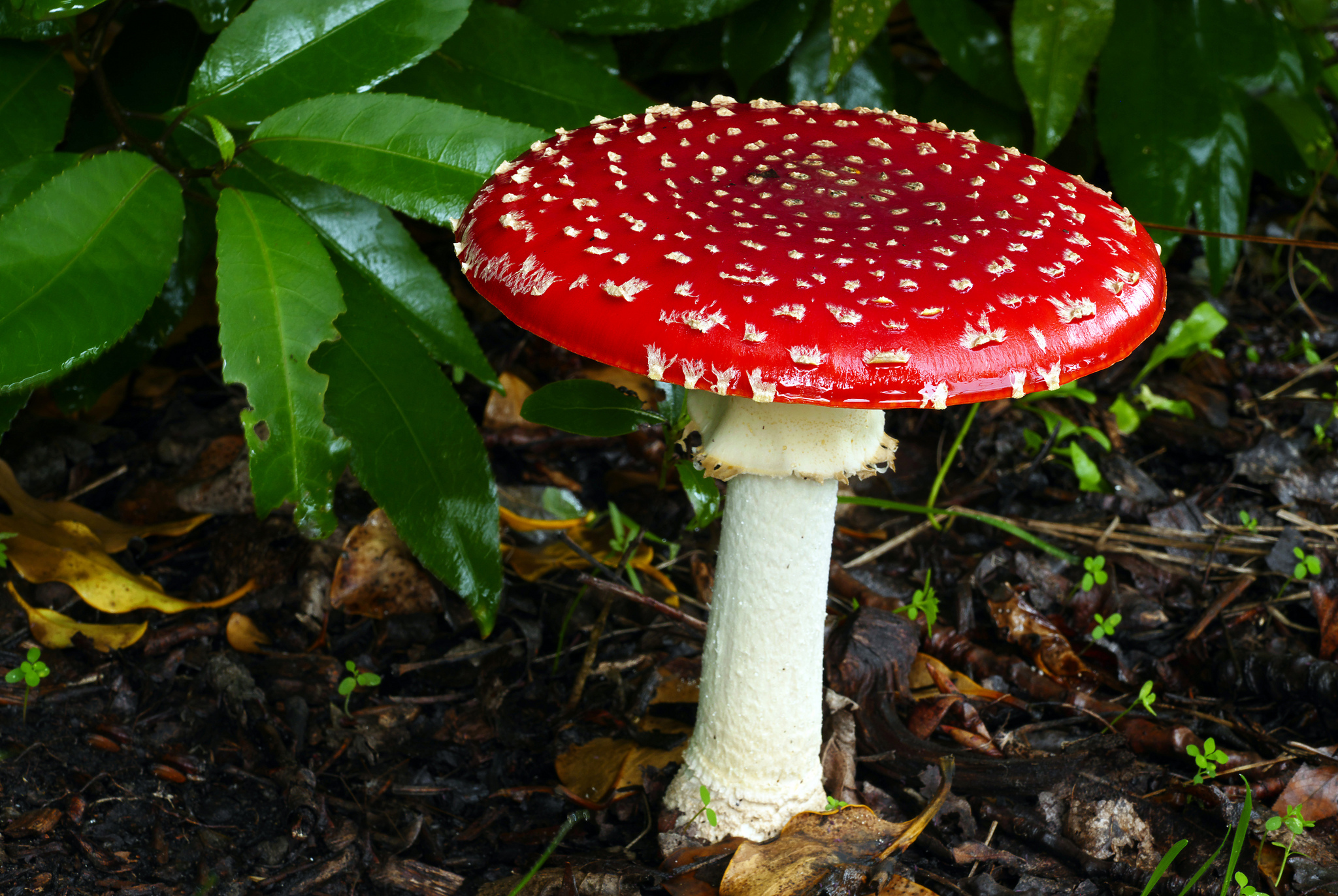}
\includegraphics[height=.30\linewidth]{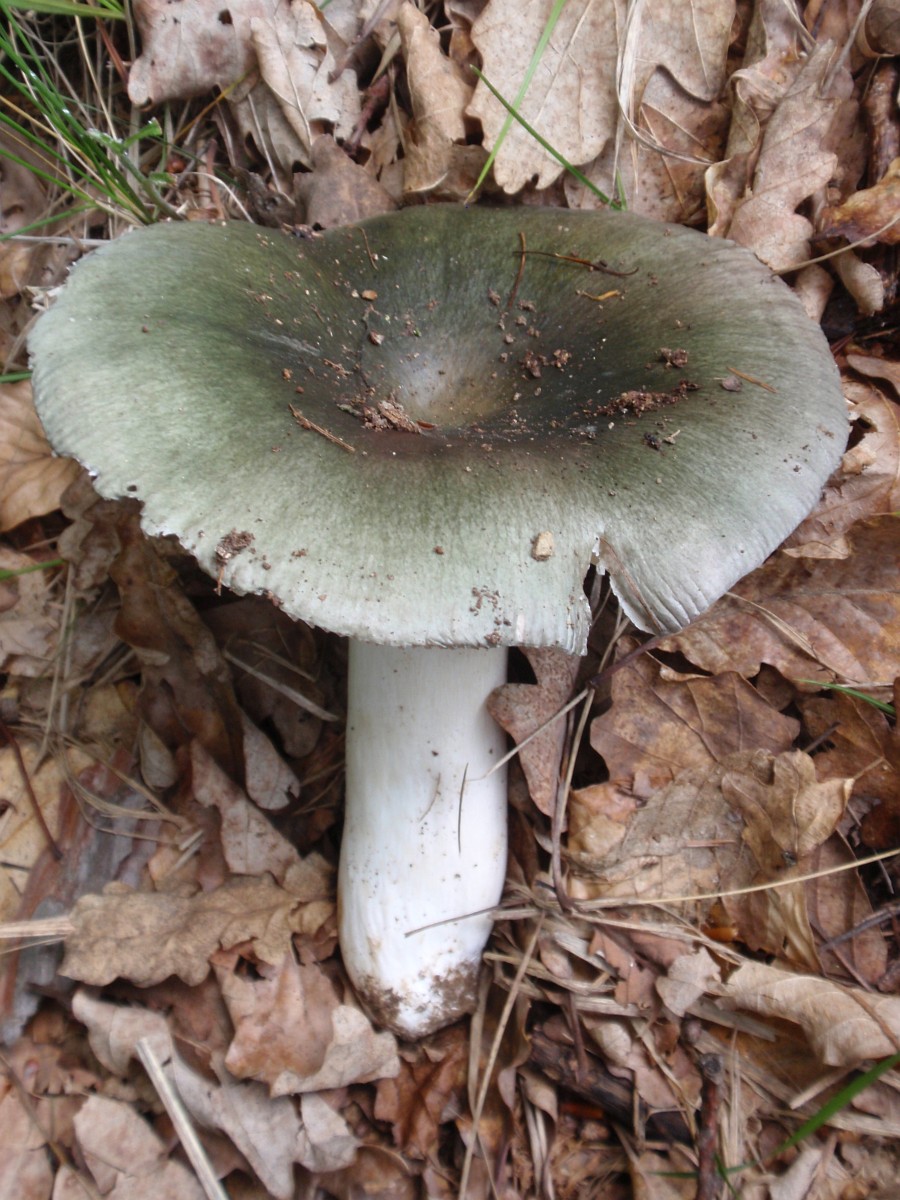}
\includegraphics[height=.30\linewidth]{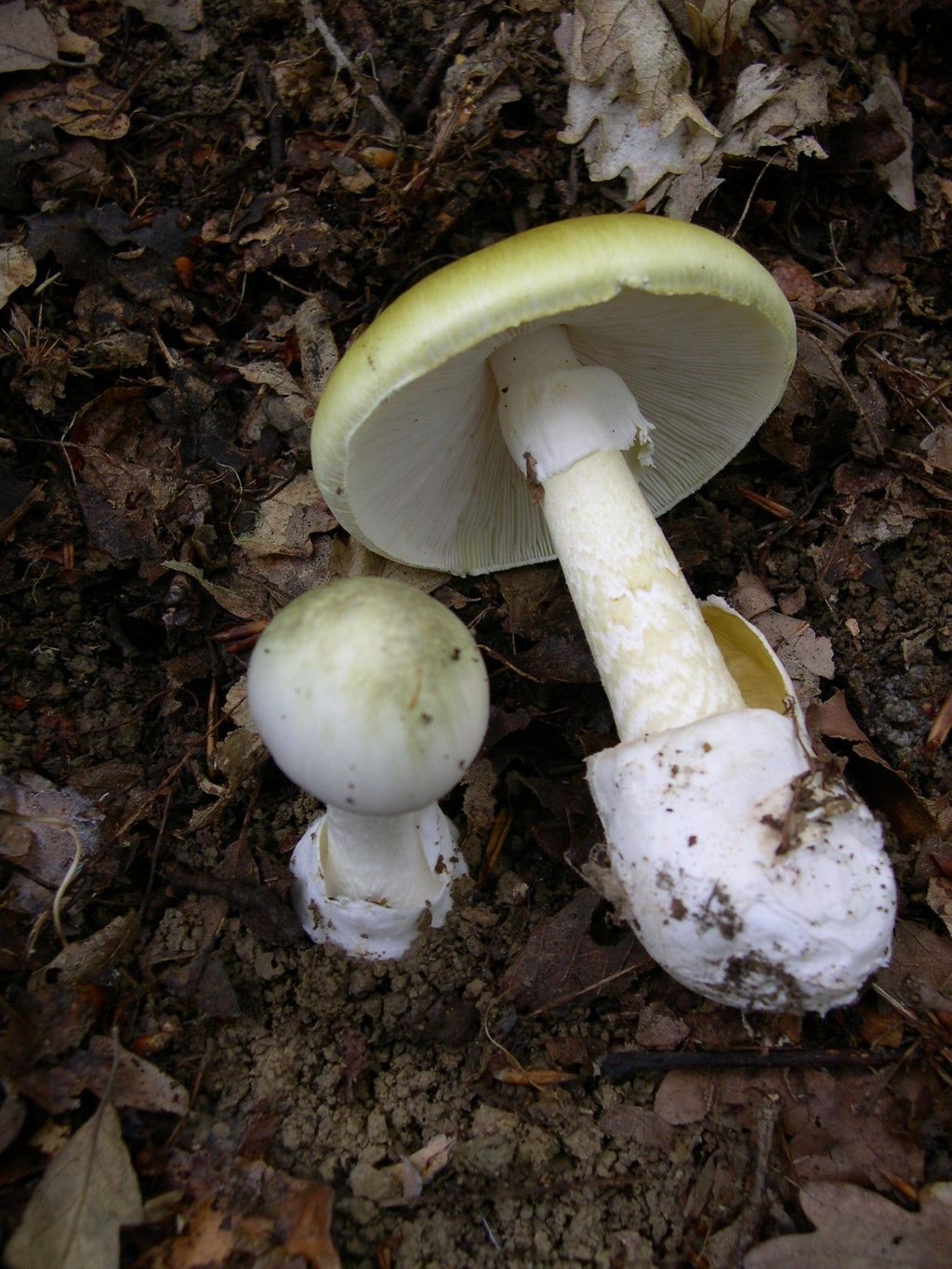}
\end{center}
\caption{One image of a mushroom (\emph{Muscaria}) may be enough to recognize it in the wild (left); in other cases, there may be more subtle differences between an edible (\emph{Russula}, shown in the center) and a deadly one (\emph{Phalloides}, shown on the right), but still few samples are enough for humans.}
\label{fig:motivation}
\end{figure}
\section{Introduction}

Consider Figure~\ref{fig:motivation}: Given one or few images of an \emph{Amanita Muscaria} (left), one can easily recognize it in the wild.
    Identifying a \emph{Russula} (center)  may require more samples, enough to distinguish it from the deadly \emph{Amanita Phalloides} (right), but likely not millions of them. We refer to this as {\em few-shot learning.} This ability comes from having seen and touched millions of other objects, in different environments, under different lighting conditions, partial occlusions and other nuisances. We refer to this as {\em meta-learning.} We wish to exploit the availability of large annotated datasets to meta-train models so they can learn new concepts from few samples, or ``shots.'' We refer to this as {\em meta-training for few-shot learning.}
 
In this paper we develop a framework for both meta-training (learning a potentially large number of classes from a large annotated dataset) and few-shot learning (using the learned model to train new concepts from few samples), designed to have the following characteristics.

\noindent{\bf Open set:} Accommodate an unknown, growing, and possibly unbounded number of new classes in an {\em ``open set'' or ``open universe''} setting. Some of the simpler methods available in the literature, for instance based on nearest-neighbors of fixed embeddings \cite{Snell:nips17}, do so in theory. In these methods, however, there is no actual few-shot {\em learning} per se, as all learnable parameters are set at meta-training.

\noindent{\bf Continual:} Enable leveraging few-shot data to improve the model parameters, even those inferred during meta-training. While each class may only have few samples, as the number of classes grows, the few-shot training set may grow large. We want a model flexible enough to enable {\em ``lifelong''} or {\em ``continual''} learning.

\noindent{\bf Shot Free:}  Accommodate a variable number of shots for each new category. Some classes may have a few samples, others a few hundred; we do not want to meta-train different models for different number of shots, nor to restrict ourselves to all new classes having the same number of shots, as many recent works do. This may be a side-effect of the benchmarks available  that only test a few combinations of shots and ``ways'' (classes). 

\noindent{\bf Embedded Class Models:} {\em Learn} a representation of the classes that is not constrained to live in the same space as the representation of the data. All known methods for few-shot learning choose an explicit function to compute class representatives (a.k.a. ``prototypes'' \cite{Snell:nips17}, ``proxies,'' ``means,'' ``modes,'' or ``templates'') as some form of averaging in the embedding (feature) space of the data. By decoupling the data (feature space) from the classes (class embedding), we free the latter to live in a richer space, where they  can better represent complex distributions, and possibly grow over time.

To this end, our contributions are described as follows:
\begin{itemize}
    \item \noindent{\bf Shot-free:} A meta-learning model and sampling scheme that is suitable for use with any number of ways and any number of shots, and can operate in an open-universe, life-long setting. When we fix the shots, as done in the benchmarks, we achieve essentially state-of-the-art performance, but with a model that is far more flexible.

\item \noindent{\bf Embedded Identities:} We abstract the identities to a different space than the features, thus enabling capturing more complex classes.

\item \noindent{\bf Implicit Class Representation:}  The class representation function has a variable number of arguments, the shots in the class. Rather than fixing the number of shots, or choosing a complex architecture to handle variable numbers, we show that learning an {\em implicit form} of the class  function enables seamless meta-training, while requiring a relatively simple optimization problem to be solved at few-shot time. We do not use either  recurrent architectures that impose artificial ordering, or complex set-functions.

\item \noindent{\bf Metric Learning} is incorporated in our model, enabling us to add new classes without crowding the class representation space. 

\item \noindent{\bf Performance:} Since there is no benchmark to showcase all the features of our model, we use existing benchmarks for few-shot learning that fix the number of ways and shots to a few samples. Some of the top performing methods are tailored to the benchmark, training different models for different number of shots, which does not scale, and does not enable handling the standard case where each way comes with its own number of shots. Our approach, while not tuned to any benchmark, achieves state-of-the-art performance and is more general.
\end{itemize}
In the next section we present a formalism for ordinary classification that, while somewhat pedantic, allows us to generalize to life-long, open universe, meta- and few-shot training. The general model allows us to analyze existing work under a common language, and highlights limitations that motivate our proposed solution in Sect. \ref{sec:related_work}.

\subsection{Background, Notation; Ordinary Classification}

In ordinary classification, we call ${\cal B} = \{(x_i, y_i) \}_{i = 1}^M$, with $y_i \in \{1, \dots, B\}$ a ``large-scale'' training set;
$(x_j, y_j) \sim P(x,y)$  a sample from the same distribution. If it is in the training set, 
we write formally 
$P(y = k | x_i) = \delta(k-y_i)$. 
Outside the training set,  we approximate this probability with
\begin{equation}
    P_w(y = k| x) := \frac{\exp(-\phi_w(x)_k)}{\sum_k \exp(-\phi_w(x)_k)}
    \label{eq:phi}
\end{equation} 
where the discriminant $\phi_w: X\rightarrow \mathbb{R}^K$ is an element of a sufficiently rich parametric class of functions with parameters, or ``weights,'' $w$, and the subscript $k$ indicates the $k$-th component. The empirical cross-entropy loss is defined as 
\begin{eqnarray}
    L(w) &:=& \sum_{\stackrel{k=1}{(x_i, y_i) \in {\cal B}}}^K  -P(y = k |x_i)\log P_w(y = k |x_i) \nonumber \\
    & = & \sum_{(x_i, y_i) \in {\cal B}} -\log P_w(y_i | x_i)
    \label{eq:L}
\end{eqnarray}
minimizing which is equivalent to maximizing $\prod_i P_w(y_i | x_i)$.  If ${\cal B}$ is i.i.d., this yields the maximum-likelihood estimate $\hat w$, 
that depends on the dataset $\cal B$ and approximates $\phi_{\hat w}(x)_y \simeq \log P(y|x)$. We write cross-entropy explicitly as a function of the discriminant as
\begin{equation}
    L(w) = \sum_{(x_i, y_i) \in {\cal B}} \ell(\phi_w(x_i)_{y_i}) 
    \label{eq:H}
\end{equation}
by substituting \eqref{eq:phi} into \eqref{eq:L}, where $\ell$ is given, with a slight abuse of notation, by
\begin{equation}
    \ell(v_i) := -v_i + {\rm LSE}(v) 
    \label{eq:ell}
\end{equation}
with the log-sum-exp ${\rm LSE}(v) := \log\left(\sum_{k=1}^K \exp( v_k) \right)$.
Next, we introduce the general form for few-shot and life-long learning, used later to taxonomize modeling choices made by different approaches in the literature.

\subsection{General Few-Shot Learning}

Let ${\cal F} = \{(x_j, y_j)\}_{j = 1}^{N(k)}$ be the few-shot training set, with $k \in \mathbb{N}$ the classes, or ``ways,'' and $N(k)$ the ``shots,'' or samples per class.  We assume that meta- and few-shot data $x_i, x_j$ live in the same domain ({\em e.g.}, natural images), while the meta- and few-shot classes are disjoint, which we indicate with $y \in B+\{1, \dots, K \}$.\footnote{The number of ways $K$ is a-priori unknown and potentially unbounded. It typically ranges from a few to few hundreds, while $N(k)$ is anywhere from one to a few thousands. The meta-training set has typically $M$ in the millions and $B$ in the thousands. Most benchmarks assume the same number of shots for each way, so there is a single number $N$, an artificial and unnecessary restriction. There is no loss of generality in assuming the classes are disjoint, as few-shot classes that are shared with the meta-training set can just be incorporated into the latter.}

During meta-training, from the dataset ${\cal B}$ we learn a parametric representation (feature, or embedding) of the data $\phi_w(x)$, for use later for few-shot training. During few-shot training, we use $N(k)$ samples for each new category $k>B$ to train a classifier, with $k$ potentially growing unbounded (life-long learning). First, we define ``useful'' and then formalize a criterion to learn the parameters $w$, both during meta- and few-shot training.

Unlike standard classification, discussed in the previous section, here we do not know the number of classes ahead of time, so we need a representation that is more general than a $K$-dimensional vector $\phi_w$. To this end, consider two additional ingredients: A representation of the classes $c_k$  (identities, prototypes, proxies), and a mechanism to associate a datum $x_j$ to a class $k$ through its representative $c_k$. We therefore have three functions, all in principle learnable and therefore indexed by parameters $w$. The {\em data representation} $\phi_w: X \rightarrow \mathbb{R}^F$ maps each datum to a fixed-dimensional vector, possibly normalized,
\begin{equation}
  z = \phi_w(x).
  \label{eq:z}
\end{equation}
We also need a {\em class representation}, that maps the $N(k)$ features $z_j$ sharing the same identity $y_j = k$, to some representative $c_k$ through a function $\psi_w: \mathbb{R}^{F N(k)} \rightarrow \mathbb{R}^C$ that yields, for each $k = B+1, \dots, B+K$
\begin{equation}
  c_k = \psi_w\left(\{z_j \ | \ y_j = k\}\right)
  \label{eq:ck}
\end{equation}
where $z_j = \phi_w(x_j)$. Note that the argument of $\psi$ has variable dimension. Finally, the {\em class membership} can be decided based on the posterior probability of a datum belonging to a class, approximated with a sufficiently rich parametric function class in the exponential family as we did for standard classification, 
\begin{equation}
    P_w(y = k | x_j) := \frac{\exp\left( - \chi_w(z_j, c_k)\right)}{\sum_k \exp(-\chi_w(z_j, c_k))}
    \label{eq:chi}
\end{equation}
where $\chi_w: \mathbb{R}^F \times \mathbb{R}^C \rightarrow \mathbb{R}$ is analogous to \eqref{eq:phi}. The cross-entropy loss \eqref{eq:L} can then be written as
\begin{equation}
     L(w) = \sum_{k = B+1}^{B+K}\sum_{j = 1}^{N(k)} \ell( \chi_w(z_j, c_k))
\end{equation}
with $\ell$ given by \eqref{eq:ell} and $c_k$ by \eqref{eq:ck}. The loss is minimized when $\chi_{\hat w}(z_j, c_k) = \log P(y_j = k | x_j),$ a function of the few-shot set ${\cal F}$. Note, however, that this loss can also be applied to the meta-training set, by changing the outer sum to $k = 1, \dots, B$, or or to any combination of the two, by selecting subsets of $\{1, \dots, B+K\}$. Different approaches to few-shot learning differ in the choice of model $\cal M$ and mixture of meta- and few-shot training sets used in one iteration of parameter update, or training ``episode.''

\section{Stratification of Few-shot Learning Models}
\label{sec:stratification}

Starting from the  most general form of few-shot learning described thus far, we restrict the model until there is no few-shot learning left, to capture the modeling choices made in the literature. 

\subsection{Meta Training}
In general, during meta-training for few-shot learning, one solves some form of
\begin{multline}
\hat w = \arg\min_w  \underbrace{\sum_{(x_i, y_i) \in {\cal B}} \ell (\chi_w(z_i, c_i))}_{L(w,c)} \\
{\rm s. \ t. \ } z_i = \phi_w(x_i); \ c_i = \psi_w(\{z_j \ | y_j = i\}).
\nonumber
\end{multline}

\noindent{\bf Implicit class representation  function:} Instead of the explicit form in \eqref{eq:ck}, one can infer the function $\psi_w$ implicitly: Let $r = \min_w L(w,\psi_w)$ be the minimum of the optimization problem above. If we consider $c = \{c_1, \dots, c_B\}$ as free parameters in $L(w, c)$, the equation $ r = L(\hat w, c)$ defines $c$ implicitly as a function of $\hat w$, $\psi_{\hat w}$. 
One can then simply find $\hat w$ and $c$ simultaneously by solving
\begin{equation}
\hat w, \hat c = \arg\min_{w,c}  \sum_{\stackrel{k = 1}{i | y_i = k} }^B 
\ell (\chi_w(\phi_w(x_i), c_k))
\label{eq:meta}
\end{equation}
which is equivalent to the previous problem, even if there is no explicit functional form for the class representation $\psi_w$. As we will see, this simplifies  meta-learning, as there is no need to design a separate architecture with a variable number of inputs $\psi_w$, but requires solving a (simple) optimization during few-shot learning. This is unlike all other known few-shot learning methods, that learn or fix $\psi_w$ during meta-learning, and keep it fixed henceforth.

Far from being a limitation, the implicit solution has several advantages, including bypassing the need to explicitly define a function with a variable number of inputs (or a set function) $\psi_w$. It also enables the identity representation to live in a different space than the data representation, again unlike existing work that assumes a simple functional form such as the mean.

\subsection{Few-shot Training}

\noindent{\bf Lifelong few-shot learning:} Once meta-training is done, one can use the same loss function in \eqref{eq:meta} for $k>B$ to achieve life-long, few-shot learning. While each new category $k>B$ is likely to have few samples $N(k)$, in the aggregate the number of samples is bound to grow beyond $M$, which we can exploit to update both the embedding $\phi_w$, the metric $\chi_w$ and the class  function $c_k = \psi_w$.

\noindent{\bf Metric learning:} A simpler model consists of fixing the parameters of the data representation $\hat \phi := \phi_{\hat w}$ and using the same loss function, but summed for $k>B$, to learn from few shots $N_k$ the new class proxies $c_k$ and change the metric $\chi_w$ as the class representation space becomes crowded. If we fix the data representation, during the few-shot training phase, we solve
\begin{equation}
\hat w, \hat c = \arg\min_{w,c} \sum_{k = B+1}^{B+K}\sum_{j | y_j = k}\ell(\chi_w(\hat \phi(x_j), c_k))
\label{eq:backfill}
\end{equation}
where the dependency on the meta-training phase is through $\hat \phi$ and both $\hat w$ and $\hat c$ depend on the few-shot dataset ${\cal F}$.

\noindent{\bf New class identities:} 
One further simplification step is to also fix the metric $\chi$, leaving only the class representatives to be estimated
\begin{equation}
    \hat c = \arg\min_c\sum_{k = B+1}^{B+K}\sum_{j | y_j = k} \ell(\chi(\hat \phi(x_j), c_k)). 
    \label{eq:c-implicit}
\end{equation}
The above is the implicit form of the parametric function $\psi_w$, with parameters $w = c$, as seen previously. Thus evaluating $\hat c_k = \psi_c(\{z_j \ | y_j = k\})$ requires solving an optimization problem.

\noindent{\bf No few-shot learning:} Finally, one can fix even the function $\psi$ explicitly, forgoing few-shot learning and simply computing
\begin{equation}
    \hat c_k = \psi(\{\hat \phi(x_j) \ | y_j  = k\}), \ k > B
    \label{eq:nolearn}
\end{equation}
that depends on ${\cal B}$ through $\hat \phi$,  and on ${\cal F}$ through $Y_k$.  

We articulate our modeling and sampling choices in the next section, after reviewing the most common approaches in the literature in light of the stratification described. 

\subsection{Related Prior Work}
\label{sec:related_work}

Most current approaches fall under the case~\eqref{eq:nolearn}, thus involving no few-shot learning, forgoing the possibility of lifelong learning and imposing additional undue limitations by constraining the prototypes to live in the same space of the features. Many are variants of Prototypical Networks \cite{Snell:nips17}, where only one of the three components of the model is learned:  $\psi$ is fixed to be the mean, so $c_k : = \frac{1}{|Y_k|}\sum_{j\in Y_k} z_j$ and $\chi(z,c) =\| z - c \|^2$ is the Euclidean distance. The only learning occurs at meta-training, and the trainable portion of the model $\phi_w$ is a conventional neural network. In addition, the sampling scheme used for training makes the model dependent on the number of shots, again unnecessarily.

Other work can be classified into two main categories: gradient based \cite{meta-lstm,MAML,SNAIL,LEO} and metric based \cite{Snell:nips17,matching-net,Oreshkin:Nips18,Gidaris:cvpr18}. In the first, a \emph{meta-learner} is trained to adapt the parameters of a network to match the few-shot training set. \cite{meta-lstm} uses the base set to learn long short-term memory (LSTM) units \cite{LSTM} that update the base classifier with the data from the few-shot training set. MAML \cite{MAML} learns an initialization for the network parameters that can be adapted by gradient descent in a few steps. LEO \cite{LEO} is similar to MAML, but uses a task specific initial condition and performs the adaptation in a lower-dimensional space. Most of these algorithms adapt $\phi_w(x)$ and use an ordinary classifier at few-shot test time. There is a different $\phi_w(x)$ for every few-shot training set, with little re-use or any continual learning.

On the metric learning side, \cite{matching-net} trains a weighted classifier using an \emph{attention mechanism} \cite{XuEtAl:AttentionMechanism:ICML:2015} that is applied to the output of a feature embedding trained on an the base set. This method requires the shots at meta- and few-shot training to match. \emph{\pns} \cite{Snell:nips17}  are trained with episodic sampling and a loss function based on the performance of a nearest-mean classifier \cite{TibshiraniEtAl:CancerDiagNearestCentroid:PNAS:2002} applied to a few-shot training set. \cite{Gidaris:cvpr18}  generates classification weights for a novel class based on a feature extractor using the base training set.
Finally, \cite{r2d2} incorporates ridge regression in an end-to-end manner into a deep-learning network.
These methods learn a single $\phi_w(x)$, which is reused across few-shot training tasks. The class identities are then either  obtained through a function defined a-priori such as the sample mean in \cite{Snell:nips17}, an attention kernel \cite{matching-net}, or ridge regression \cite{r2d2}. The form of $\psi_w$ or $\chi$ do not change at few-shot training. \cite{Oreshkin:Nips18}  uses task-specific adaptation networks to facilitate the adapting embedding network with output on a task-dependent metric space. In this method, the form of $\chi$ and $\psi$ are fixed and the output of $\phi$ is modulated based on the few-shot training set.

Next, we describe our model that, to the best of our knowledge, is the first and only to learn each component of the model: The embedding $\phi_w$, the metric $\chi_w$, and implicitly the class representation $\phi_w$.


\section{Proposed Model} 
Using the formalism of Sect.~\ref{sec:stratification} we describe our modeling choices. Note that there is redundancy in the model class ${\cal M}$, as one could fix the data representation $\phi(x) = x$, and devolve all modeling capacity to $\psi$, or vice-versa. The choice depends on the application context. We outline our choices, motivated by limitations of prior work. 


\noindent{\bf Embedding} $\phi_w$: In line with recent work, we choose a deep convolutional network. The details of the architecture are in Sect. \ref{sec:implementation}.

\noindent{\bf Class representation function} $\psi_w$: We define it implicitly by treating the class representations $c_k$ as parameters along with the weights $w$. As we saw earlier, this means that at few-shot training, we have to solve a simple optimization problem \eqref{eq:c-implicit} to find the representatives of new classes, rather than computing the mean as in Prototypical Networks and its variants:
\begin{equation}
    c_k = \arg\min_c \sum_{j | y_j = k} \ell(\chi_w(\hat \phi(x_j), c)) = \psi_c(k).
\end{equation}
Note that the class estimates depend on the parameters $w$ in $\chi$. If few-shot learning is resource constrained, one can still learn the class representations implicitly during meta-training, and approximate them with a fixed function, such as the mean, during the few-shot phase.

\noindent{\bf Metric} $\chi$: we choose a discriminant induced by the Euclidean distance in the space of class representations, to which data representations are mapped by a learnable parameter matrix $W$:
\begin{equation}
    \chi_{{}_W}(z_j, c_k) = \| W \hat \phi(x_j)-  c_k \|^2
    \label{eq:metric-learning-linear}
\end{equation}
Generally, we pick the dimension of $c$ larger than the dimension of $z$, to enable capturing complex multi-modal identity representations. Note that this choice encompasses metric learning: If $Q = Q^T$ was a symmetric matrix representing a change of inner product, then $\|W \phi - c\|^2_Q = \phi^T W^T Q c$ would be captured by simply choosing the weights $\tilde W  = QW$. Since both the weights and the class proxies as free, there is no gain in generality in adding the metric parameters $Q$. Of course, $W$ can be replaced by any non-linear map, effectively ``growing'' the model via
\begin{equation}
    \chi_{w}(z_j, c_k) = \|  \hat f_w(\phi(x_j)) - c_k \|^2
\end{equation}
for some parametric family $f_w$ such as a deep neural network.

\section{Implementation}
\label{sec:implementation}
\paragraph{Embedding  $\phi_w(x_j)$} We use two different architectures. The first  \cite{Snell:nips17,matching-net} is four-convolution blocks, each block with 64 $3\times3$ filters followed by batch-normalization and ReLU. This is passed through max-pooling of a $2\times 2$ kernel. Following the convention in \cite{Gidaris:cvpr18}, we call this architecture C64. The other network is a modified ResNet \cite{resnet},  similar to \cite{Oreshkin:Nips18}. We call this ResNet-12. 

In addition, we normalize the embedding to live on the unit sphere, \ie $\phi(x) \in \mathbb{S}^{d-1}$, where $d$ is the dimension of the embedding. This normalization is added as a layer to ensure that the feature embedding are on the unit sphere, as opposed to applying it post-hoc.
This adds some complications during meta-training due to poor scaling of gradients \cite{Wang:2017}, and is addressed by a single parameter layer after normalization, whose sole purpose is scaling the output of the normalization layer. This layer is not required at test time.

\paragraph{Class representation:} As noted earlier, this is implicit during meta-training. 
In order to show the flexibility of our framework, we increase the dimension of the class representation. 

\paragraph{Metric $\chi$} We choose the angular distance in feature space, which is the $d$-hypersphere: 
\begin{align}
\chi(z_j,c_k) = \|W z_j-  c_k\|^{2} &= 2s^{2}(1 - \cos\theta),
\end{align}
where $s$ is the scaling factor used during training and $\theta$ the angle between the normalized arguments.  As the representation $z = \phi_w(x)$ is normalized, the class-conditional model is a Fisher-Von Mises (spherical Gaussian). However, as $W\phi_w(x_i) \in \mathbb{S}^{d-1}$, we need $W \psi_w \in \mathbb{S}^{d-1}$. 
During meta-training we apply the same normalization and scale function to the implicit representation as well.
\begin{equation}
    P_w(y=k|x) \propto \exp \langle W \phi_w(x), c_{k} \rangle
\end{equation}
up to the normalization constant.

\paragraph{Sampling}
At each iteration during meta-training, images from the training set $\mathcal{B}$ are presented to the network in the form of \emph{episodes} \cite{matching-net,meta-lstm,Snell:nips17}; each episode consists of images sampled from $K$ classes. The images are selected by first sampling $K$ classes from $\mathcal{B}$ and then sampling $N_{e}$ images from each of the sampled classes. The loss function is now restricted to the $K$ classes present in the episode as opposed to the entire set of classes available at meta-training. This setting allows for the network to learn a better embedding for an open set classification as shown in \cite{closerlook,matching-net} 

Unlike existing sampling methods that use episodic sampling~\cite{meta-lstm,Snell:nips17}, we do not split the images within an episode into a meta-train set and a meta-test set. For instance, prototypical networks \cite{Snell:nips17} use the elements in the meta-train set to learn the mean of the class representation. \cite{meta-lstm} learns the initial conditions for optimization. This requires a notion of training ``shot,'' and results in multiple networks to match the shots one expects at few-shot training. 

\paragraph{Regularization}
First, we notice that the loss function  \eqref{eq:meta} has a degenerate solution where all the centers and the embeddings are the same. In this case,  $P_w(y=k|x_j) = P_w(y=k'|x_j)$ for all $k$ and $k'$, \ie, $P_w(y=k'|x_j)$ is a uniform distribution. For this degenerate case, the entropy is maximum, so we use entropy to bias the solution away from the trivial one. We also use Dropout \cite{dropout} on top of the embedding $\phi_w(x)$ during meta-training. Even when using episodic sampling, the embedding tends to over-fit on the base set in the absence of dropout. We do not use this at few-shot train and test time.   

\begin{figure}
    \centering
    \includegraphics[width=0.99\linewidth]{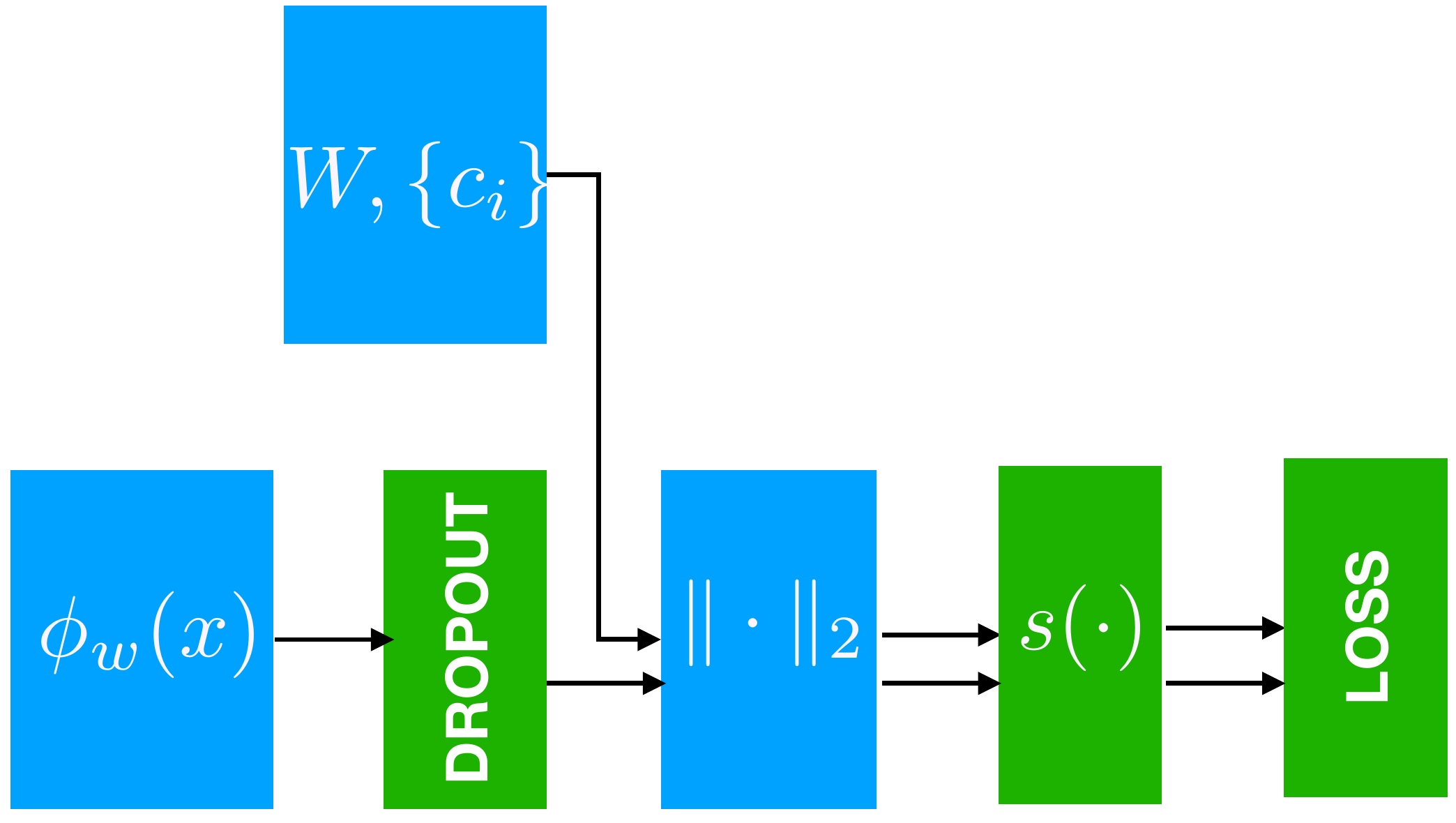}
    \caption{Our meta-training loss flow: The layers represented in blue are the layers that remain after meta-training. While the green layers  are used only for training. Here $\|\cdot\|$ represents an $L_2$ normalization layer and $s(\cdot)$ represents a scaling layer}
    \label{fig:my_label}
\end{figure}

\figref{fig:my_label} summarizes our architecture for the loss function during meta training. This has layers that are only needed for training such as the scale layer, Dropout and the loss. During few-shot training, we only use the learned embedding $\phi_w(x).$


\section {Experimental Results}
We test our algorithm on three datasets: \mi{} \cite{matching-net}, \ti{} \cite{Ren:2018} and \fs{}  \cite{r2d2}. The \mi dataset consists of images of size $84 \times 84$ sampled from 100 classes of the ILSVRC \cite{imagenet} dataset, with 600 images per class.
We used the data split outlined in \cite{meta-lstm}, where 64 classes are used for training, 16 classes are used for validation, and 20 classes are used for testing.

We also use \ti{}  \cite{Ren:2018}.
This is a larger subset of ILSVRC, and consists of 779,165 images of size $84 \times 84$ representing 608 classes hierarchically grouped into 34 high-level classes.
The split of this dataset ensures that sub-classes of the 34 high-level classes are not spread over the training, validation and testing sets, minimizing the semantic overlap between training and test sets.
The result is 448,695 images in 351 classes for training, 124,261 images in 97 classes for validation, and 206,209 images in 160 classes for testing.
For a fair comparison, we use the same training, validation and testing splits as in \cite{Ren:2018}, and use the classes at the lowest level of the hierarchy.

Finally, we use  \fs{}, (CIFAR-FS) \cite{r2d2} containing images of size $32\times32$, a reorganized version of the CIFAR-100 \cite{cifar100} dataset.
We use the same data split as in \cite{r2d2}, dividing the 100 classes into 64 for training, 16 for validation, and 20 for testing. 

\subsection{Comparison to \Pns}
Many recent methods are variants of \pns, so we perform detailed comparison with it. We keep the training procedure, network architecture, batch-size as well as data augmentation the same. The performance gains are therefore solely due to the improvements in our method. 

We use ADAM~\cite{Kingma2015Adam:Optimization} for training with an initial learning rate of $10^{-3}$, and a decay factor of $0.5$  every 2,000 iterations. We use the validation set to determine the best model. Our data augmentation consists of mean subtraction, standard-deviation normalization, random cropping and random flipping during training. Each episode contains 15 query samples per class during training. In all our experiments, we set $\lambda=1$ and did not tune this parameter. 

Except otherwise noted, we always test few-shot algorithms on 2000 episodes, with 30 query classes per point per episode. At few-shot training, we experimented with setting the class identity to be  implicit (optimized) or average prototype (fixed). The latter may be warranted when the few-shot phase is resource-constrained and yields similar performance. To compare computation time, we use the fixed mean. Note that, in all cases, the class prototypes are learned implicitly during meta-training.  

The results of this comparison are shown in \tabref{tab:baseline}. From this table we see that for the 5-shot 5-way case we perform similarly to \pn. However, for the 1-shot case we see significant improvements across all three datasets. Also, the performance of \pns drops when the train and test shot are changed. \tabref{tab:baseline} shows a significant drop in performance when we test models with a 5-shot setting and train with 1-shot. Notice that, from the table, our method is able to maintain the same  performance. Consequently,  we only train \textbf{one} model and test it across the different shot scenarios, hence the moniker ``shot-free.''

\begin{table*}[htbp]
  \begin{center}
  \begin{tabular}{@{}|c|c|c|c|c|c @{}}
    \hline
    Dataset & Testing Scenario & Training Scenario &   Our implementation of  \cite{Snell:nips17} & Our Method \\ 
        \hline
    \multirow{ 3}{*}{\mi} & \shot{1}{5} & \shot{1}{5} &   43.88 $\pm$ 0.40 & \textbf{49.07 $\pm$ 0.43} \\ 
      & \shot{5}{5} &  \shot{1}{5}  & 58.33 $\pm$ 0.35 & \textbf{64.98 $\pm$ 0.35}   \\ 
      & \shot{5}{5} &  \shot{5}{5} & 65.49  $\pm$ 0.35 & \textbf{65.73 $\pm$ 0.36}   \\ 
        \hline
            \hline
    \multirow{ 3}{*}{\ti} &\shot{1}{5} & \shot{1}{5}  & 41.36 $\pm$ 0.40 & \textbf{48.19 $\pm$ 0.43}\\ 
    & \shot{5}{5} &  \shot{1}{5} & 55.93 $\pm$ 0.39 & \textbf{64.60 $\pm$ 0.39}\\ 
     & \shot{5}{5} &  \shot{5}{5} & 65.51 $\pm$ 0.38 & 65.50 $\pm$ 0.39\\ 
        \hline
             \hline
    \multirow{ 3}{*}{\fs} &\shot{1}{5} & \shot{1}{5}  &  50.74 $\pm$ 0.48 & \textbf{55.14 $\pm$ 0.48}\\ 
    &\shot{5}{5} &  \shot{1}{5} & 64.63 $\pm$ 0.42 & \textbf{70.33 $\pm$ 0.40}\\ 
    &\shot{5}{5} &  \shot{5}{5} & 71.57 $\pm$ 0.38 & \textbf{2x $\pm$ 0.39}\\ 
        \hline
  \end{tabular}
  \end{center}
  \caption{Comparison of results from our method to that of our implementation of \pn{} \cite{Snell:nips17} using the C64 network architecture.
  The table shows the accuracy and 95\% percentile confidence interval of our method averaged over 2,000 episodes on different datasets. Note that our method does not have a notion of shot, here we when we imply training by different shot, we mean that the batch sizes is the same as that of the prescribed method.}
 \label{tab:baseline}
\end{table*}

\subsection {Effect of Dimension of Class Identities}
Class identities $c_k$ can live in a space of different dimensions than the feature embedding. This can be done in two ways: by lifting the embedding into a higher dimension space or by projecting the class identity into the embedding dimension.  If the dimension of the class identity changes, we also need to modify $\chi$ according to \eqref{eq:metric-learning-linear}.  The weight matrix $W \in \mathbb{R}^{d\times \mu}$, where $d$ is the dimension of the embedding and $\mu$ is the dimension of the class identities, can be learned during meta-training. This is equivalent to adding a fully connected layer through which the class identities are passed  before  normalization. Thus, we now learn $\phi_w$, $\psi_k$ and $\chi_W$. We show experimental results with the $C64$ architecture on the \mi datasets in \tabref{tab:meandim}. Here, we tested the dimension of the class identities to be  $2\times$,~$5\times$ and $10\times$ the dimension of the embedding. From this table we see that increasing the dimensions gives us a performance boost. However, this increase saturates at a dimension of $2\times$ the dimension of the embedding space.

\begin{table}[htbp]
  \begin{center}
  \resizebox{\textwidth}{!}{
  \begin{tabular}{@{} |c|c|c|c|c| @{}}
    \hline
     Dimension &  1x  &  2x    & 5x & 10x \\ 
    \hline
     Performance &  49.07  & 51.46   & 51.46   & 51.32\\
    \hline
  \end{tabular}}
  \end{center}
  \caption{Performance of our method on \mi  with the class identity dimension as a function of the embedding dimension using the C64 network architecture. The table shows the accuracy averaged over 2,000 episodes.}
  \label{tab:meandim}
\end{table}

\subsection{Comparison to the State-of-the-art}
In order to compare with the state-of-the-art, we use the ResNet-12 base architecture, train our approach using SGD with Nesterov momentum with an initial learning rate of $0.1$, weight decay of $5e-4$, momentum of $0.9$ and eight episodes per batch. Our learning rate was decreased by a factor of $0.5$ every time the validation error did not improve for 1000 iterations. We did not tune these parameters based on the dataset. As mentioned earlier, we train \textbf{one} model and test across various shots. We also compare our method with class identities in a space with twice the dimension of the embedding. Lastly, we compare our method with  a variant of ResNet where we change the filter sizes to (64,160,320,640) from (64,128,256,512). 

The results of our comparison for \mi is  shown in \tabref{tab:misota}. Modulo empirical fluctuations, our method performs at the state-of-the art and in some cases exceeds it. We wish to point out that SNAIL \cite{SNAIL}, TADAM  \cite{Oreshkin:Nips18,MTLF}, LEO \cite{LEO}, MTLF \cite{MTLF} pre-train the network for a 64 way classification task on \mi and 351 way classification on \ti. However, all the models trained for our method are trained from scratch and use no form of pre-training. We also do not use the meta-validation set for tuning any parameters other than selecting the best trained model using the error on this set. Furthermore, unlike all other methods, we did not have to train multiple networks and tune the training strategy for each case. Lastly, LEO \cite{LEO} uses a very deep 28 layer Wide-ResNet as a base model compared to our shallower ResNet-12. A fair comparison would involve training our methods with the same base network. However, we include this comparison for complete transparency.

\begin{table}[htbp]
  \begin{center}
  \resizebox{\textwidth}{!}{
  \begin{tabular}{@{} |c|c|c|c| @{}}
    \hline
    Algorithm &  1-shot  &  5-Shot    & 10-shot \\ 
    & 5-way & 5-way & 5-way \\
    \hline
        Meta LSTM \cite{meta-lstm} &  43.44& 60.60 &- \\
   Matching networks \cite{matching-net} &   44.20  & 57.0 &-\\ 
    MAML \cite{MAML} &   48.70 & 63.1 &-\\
    \Pns \cite{Snell:nips17} &  49.40 & 68.2 & -\\
    Relation Net \cite{relation-net} &  50.40& 65.3 & - \\
    R2D2 \cite{r2d2} & 51.20 & 68.2 &- \\
    SNAIL \cite{SNAIL} &  55.70 & 68.9 &-\\
    Gidaris\etal \cite{Gidaris:cvpr18} &  55.95 & 73.00 & -\\
TADAM \cite{Oreshkin:Nips18} &  58.50 & 76.7  & 80.8\\ 
MTFL \cite{MTLF}& 61.2 & 75.5 &- \\
LEO \cite{LEO} & 61.76 & 77.59 & -\\ 
    \hline
    Our Method (ResNet-12) &59.00    & 77.46  & 82.33\\
    Our Method (ResNet-12)  2x dims.  & 60.64    & 77.02  & 80.80 \\
    Our Method (ResNet-12 Variant)  & 59.04   & \textbf{77.64}  & \textbf{82.48} \\
    Our Method (ResNet-12 Variant)  2x dims &  60.71  & {77.26}  &  81.34 \\
    \hline
  \end{tabular}}
  \end{center}
  \caption{Performance of 4 variants of our method on \mi compared to the state-of-the-art.
  The table shows the accuracy averaged over 2,000 episodes.}
  \label{tab:misota}
\end{table}

\begin{table}[htbp]
  \begin{center}
  \resizebox{\textwidth}{!}{
  \begin{tabular}{@{} |c|c|c|c| @{}}
    \hline
    Algorithm &  1-shot  &  5-Shot    & 10-shot \\ 
    & 5-way & 5-way & 5-way \\
    \hline 
    \multicolumn{4}{c} {\ti} \\
    \hline
    MAML \cite{MAML} &   51.67 & 70.30 & - \\
    \Pns \cite{Ren:2018}  &  53.31 & 72.69 & - \\
    Relation Net \cite{relation-net} &  54.48 & 71.32 & -\\
    LEO \cite{LEO} & 65.71 & 81.31 & - \\
    \hline
    Our Method (ResNet-12) & 63.99   & 81.97 &  85.89  \\
    Our Method (ResNet-12)  2x dims.  & \textbf{66.87} & \textbf{82.64}    & 85.53 \\
    Our Method (ResNet-12)  Variant  & 63.52  & 82.59   & \textbf{86.62}   \\
    Our Method (ResNet-12)  Variant  2x dims &  \textbf{66.87}  & {82.43}  & 85.74 \\
    \hline
   \multicolumn{4}{c} {\fs} \\
    \hline
    MAML \cite{MAML} &   58.9 & 71.5 & -\\
    \Pns \cite{Snell:nips17}  &  55.5 & 72.0 & - \\
    Relation Net &  55.0 & 69.3  & - \\
    R2D2 \cite{r2d2} & 65.3 & 79.4  & - \\
    \hline
    Our Method (ResNet-12) & 69.15   &84.70 & 87.64 \\
    \hline
  \end{tabular}}
  \end{center}
  \caption{Performance of our method on \ti~and \fs datasets as compared to the state-of-the-art. The performance numbers for \fs are from \cite{r2d2}. 
  The table shows the accuracy averaged over 2,000 episodes.
  Note that the training setting for the prior work is different.}
  \label{tab:fssota}
\end{table}

The performance of our method on \ti  is shown in \tabref{tab:fssota}. This table shows that we are the top performing method for 1-shot 5-way and 5-shot 5-way. We test on this dataset as it is much larger and does not have semantic overlap between meta training and few-shot training even though fewer baselines exist for this dataset compared to \mi. Also shown in \tabref{tab:fssota} is the performance of our method on the \fs dataset. We show results on this dataset to illustrate that our method can generalize across datasets. From this table we see that our method performs the best for \fs.

\subsection{Effect of Choices in Training}
As a final remark, there is no consensus on the few-shot training and testing paradigm in the literature. There are too many variables that can affect performance. To illustrate this, we show the effect of few training choices.

\paragraph{Effect of Optimization algorithm} 
In the original implementation of \pns \cite{Snell:nips17}, ADAM~\cite{Kingma2015Adam:Optimization} was used as the optimization algorithm. However, most newer algorithms such as \cite{Oreshkin:Nips18,Gidaris:cvpr18} use SGD as their optimization algorithm. This result of using different optimization algorithms is shown in \tabref{tab:choices}. Here, we show the performance of our algorithm on the \mi dataset using a ResNet-12 model. From this table we see that, while for the \shot{1}{5} the results are better with ADAM as opposed to SGD, we see that the same does not hold for the \shot{5}{5} and \shot{10}{5}
scenarios. This shows that SGD generalizes better for our algorithm as compared to ADAM.

\begin{table}[htbp]
  \begin{center}
  \resizebox{\textwidth}{!}{
  \begin{tabular}{@{} |c|c|c|c| @{}}
    \hline
    Optimization Algorithm &  1-shot  &  5-Shot  & 10-shot   \\ 
    & 5-way & 5-way & 5-way\\
    \hline 
    ADAM &   \textbf{59.41} & 76.75 & 81.33 \\
   SGD  &  59.00 & \textbf{77.46} & \textbf{82.33} \\
    \hline
  \end{tabular}}
  \end{center}
  \caption{Performance of our method on \mi using the ResNet-12 model with different choices of optimization algorithm. The table shows the accuracy averaged over 2,000 episodes.}
  \label{tab:choices}
\end{table}

\paragraph{Effect of number of tasks per iteration.} 
TADAM \cite{Oreshkin:Nips18} and Gidaris \etal \cite{Gidaris:cvpr18} use multiple episodes per iteration. They refer to this as tasks in TADAM \cite{Oreshkin:Nips18}, which uses 2 tasks for 5-shot, 1 task for 10-shot and 5 task for 1-shot. We did not perform any such tuning and instead  defaulted it to 8 episodes per iteration based on Gidaris \etal \cite{Gidaris:cvpr18}. We also experimented with 16 episodes per iteration. However, this led to a loss in performance across all testing scenarios. \tabref{tab:choicestask}, shows the performance numbers on \mi dataset using the ResNet-12 architecture and trained using ADAM~\cite{Kingma2015Adam:Optimization} as the optimization algorithm. From this table we see that for all the scenarios 8 episodes per iteration has a better performance.

\begin{table}[htbp]
  \begin{center}
  \resizebox{\textwidth}{!}{
  \begin{tabular}{@{} |c|c|c|c| @{}}
    \hline
    Choice &  1-shot  &  5-Shot  & 10-shot   \\ 
    & 5-way & 5-way & 5-way\\
    \hline 
    8 episodes per iteration  &   \textbf{59.41} & \textbf{76.75} & \textbf{81.33} \\
    16 episodes per iteration  &  58.22 & 74.53 & 78.61 \\
    \hline
  \end{tabular}}
  \end{center}
  \caption{Performance of our method on \mi using a ResNet-12 model  with different choices of episodes per iteration. The table shows the accuracy averaged over 2,000 episodes.}
  \label{tab:choicestask}
\end{table}

 Even with all major factors such as network architecture, training procedure, batch size  remaining the same, factors such as  the number of query points used for testing these methods affect the performance and methods in  existing literature uses anywhere between 15-30 points for testing, and for some methods it is unclear what this choice was. This calls for stricter protocols for evaluation, and richer benchmark datasets.

\section{Discussion}

We have presented a method for meta-learning for few-shot learning where all three ingredients of the problem are learned: The representation of the data $\phi_w$, the representation of the classes $\psi_c$,  and the metric or membership function $\chi_W$. The method has several advantages compared to prior approaches. First, by allowing the class representation and the data representation spaces to be different, we can allocate more representative power to the class prototypes. Second, by learning the class models implicitly we can handle a variable number of shots without having to resort to complex architectures, or worse, training different architectures, one for each number of shots. Finally, by learning the membership function we implicitly learn the metric, which allows class prototypes to redistribute during few-shot learning. 

While some of these benefits are not immediately evident due to limited benchmarks, the improved generality allows our model to extend to a continual learning setting where the number of new classes grows over time, and is flexible in allowing each new class to come with its own number of shots. Our model is simpler than some of the top performing ones in the benchmarks. A single model performs on-par or better in the few-shot setting and offers added generality.

\bibliographystyle{ieee_fullname}
\balance
\bibliography{proto}

\begin{thebibliography}{10}\itemsep=-1pt

\bibitem{r2d2}
Luca Bertinetto, Jo{\~{a}}o~F. Henriques, Philip H.~S. Torr, and Andrea
  Vedaldi.
\newblock Meta-learning with differentiable closed-form solvers.
\newblock {\em CoRR}, abs/1805.08136, 2018.

\bibitem{closerlook}
Wei-Yu Chen, Yen-Cheng Liu, Zsolt Kira, Yu-Chiang~Frank Wang, and Jia-Bin
  Huang.
\newblock A closer look at few-shot classification.
\newblock In {\em International Conference on Learning Representations}, 2019.

\bibitem{MAML}
Chelsea Finn, Pieter Abbeel, and Sergey Levine.
\newblock Model-agnostic meta-learning for fast adaptation of deep networks.
\newblock In {\em ICML}, 2017.

\bibitem{Gidaris:cvpr18}
Spyros Gidaris and Nikos Komodakis.
\newblock Dynamic few-shot visual learning without forgetting.
\newblock In {\em CVPR}, 2018.

\bibitem{resnet}
Kaiming He, Xiangyu Zhang, Shaoqing Ren, and Jian Sun.
\newblock Deep residual learning for image recognition.
\newblock In {\em {CVPR}}, pages 770--778. {IEEE} Computer Society, 2016.

\bibitem{LSTM}
Sepp Hochreiter and J\"{u}rgen Schmidhuber.
\newblock Long short-term memory.
\newblock {\em Neural Comput.}, 9(8):1735--1780, Nov. 1997.

\bibitem{Kingma2015Adam:Optimization}
Diederik~P. Kingma and Jimmy~Lei Ba.
\newblock {ADAM}: A method for stochastic optimization.
\newblock {\em International Conference on Learning Representations 2015},
  2015.

\bibitem{cifar100}
Alex Krizhevsky.
\newblock Learning multiple layers of features from tiny images.
\newblock Technical report, University of Toronto, 2009.

\bibitem{SNAIL}
Nikhil Mishra, Mostafa Rohaninejad, Xi Chen, and Pieter Abbeel.
\newblock A simple neural attentive meta-learner.
\newblock In {\em ICLR}, 2018.

\bibitem{Oreshkin:Nips18}
Boris~N. Oreshkin, Pau Rodr{\'{\i}}guez, and Alexandre Lacoste.
\newblock Improved few-shot learning with task conditioning and metric scaling.
\newblock In {\em NIPS}, 2018.

\bibitem{meta-lstm}
Sachin Ravi and Hugo Larochelle.
\newblock Optimization as a model for few-shot learning.
\newblock In {\em ICLR}, 2017.

\bibitem{Ren:2018}
Mengye Ren, Eleni Triantafillou, Sachin Ravi, Jake Snell, Kevin Swersky,
  Joshua~B. Tenenbaum, Hugo Larochelle, and Richard~S. Zemel.
\newblock Meta-learning for semi-supervised few-shot classification.
\newblock {\em CoRR}, abs/1803.00676, 2018.

\bibitem{imagenet}
Olga Russakovsky, Jia Deng, Hao Su, Jonathan Krause, Sanjeev Satheesh, Sean Ma,
  Zhiheng Huang, Andrej Karpathy, Aditya Khosla, Michael Bernstein,
  Alexander~C. Berg, and Li Fei-Fei.
\newblock Imagenet large scale visual recognition challenge.
\newblock {\em Int. J. Comput. Vision}, 115(3):211--252, Dec. 2015.

\bibitem{LEO}
Andrei~A. Rusu, Dushyant Rao, Jakub Sygnowski, Oriol Vinyals, Razvan Pascanu,
  Simon Osindero, and Raia Hadsell.
\newblock Meta-learning with latent embedding optimization.
\newblock {\em CoRR}, abs/1807.05960, 2018.

\bibitem{Snell:nips17}
Jake Snell, Kevin Swersky, and Richard~S. Zemel.
\newblock Prototypical networks for few-shot learning.
\newblock In {\em {NIPS}}, pages 4080--4090, 2017.

\bibitem{dropout}
Nitish Srivastava, Geoffrey Hinton, Alex Krizhevsky, Ilya Sutskever, and Ruslan
  Salakhutdinov.
\newblock Dropout: A simple way to prevent neural networks from overfitting.
\newblock {\em J. Mach. Learn. Res.}, 15(1):1929--1958, Jan. 2014.

\bibitem{MTLF}
Qianru Sun, Yaoyao Liu, Tat{-}Seng Chua, and Bernt Schiele.
\newblock Meta-transfer learning for few-shot learning.
\newblock {\em CoRR}, abs/1812.02391, 2018.

\bibitem{relation-net}
Flood Sung, Yongxin Yang, Li Zhang, Tao Xiang, Philip~H.S. Torr, and Timothy~M.
  Hospedales.
\newblock Learning to compare: Relation network for few-shot learning.
\newblock In {\em The IEEE Conference on Computer Vision and Pattern
  Recognition (CVPR)}, June 2018.

\bibitem{TibshiraniEtAl:CancerDiagNearestCentroid:PNAS:2002}
Robert Tibshirani, Trevor Hastie, Balasubramanian Narasimhan, and Gilbert Chu.
\newblock Diagnosis of multiple cancer types by shrunken centroids of gene
  expression.
\newblock {\em Proceedings of the National Academy of Sciences},
  99(10):6567--6572, 2002.

\bibitem{matching-net}
Oriol Vinyals, Charles Blundell, Timothy Lillicrap, Koray Kavukcuoglu, and Daan
  Wierstra.
\newblock Matching networks for one shot learning.
\newblock In {\em NIPS}, 2016.

\bibitem{Wang:2017}
Feng Wang, Xiang Xiang, Jian Cheng, and Alan~Loddon Yuille.
\newblock Normface: L2 hypersphere embedding for face verification.
\newblock In {\em Proceedings of the 25th ACM International Conference on
  Multimedia}, MM '17, pages 1041--1049, New York, NY, USA, 2017. ACM.

\bibitem{XuEtAl:AttentionMechanism:ICML:2015}
Kelvin Xu, Jimmy~Lei Ba, Ryan Kiros, Kyunghyun Cho, Aaron Courville, Ruslan
  Salakhutdinov, Richard~S. Zemel, and Yoshua Bengio.
\newblock Show, attend and tell: Neural image caption generation with visual
  attention.
\newblock In {\em {ICML}}, pages 2048--2057, 2015.

\end{thebibliography}

\end{document}